\documentclass{bioinfo}
\copyrightyear{2015} \pubyear{2015}

\access{Advance Access Publication Date: Day Month Year}
\appnotes{Manuscript Category}

\usepackage{enumerate}
\usepackage{paralist}
\usepackage{url}
\usepackage[T1]{fontenc}

\begin{document}
\firstpage{1}

\subtitle{Data and text mining}

\title[The Russian Drug Reaction Corpus (RuDReC)]{The Russian Drug Reaction Corpus and Neural Models for Drug Reactions and Effectiveness Detection in User Reviews}
\author[Tutubalina \textit{et~al}.]{Elena Tutubalina\,$^{\text{\sfb 1,2}*}$, Ilseyar Alimova\,$^{\text{\sfb 1}*}$, Zulfat Miftahutdinov\,$^{\text{\sfb 1}}$, Andrey Sakhovskiy$^{\text{\sfb 1}}$, Valentin Malykh$^{\text{\sfb 1}}$, Sergey Nikolenko$^{\text{\sfb 1,2}}$}
\address{$^{\text{\sf 1}}$Kazan Federal University, 18 Kremlyovskaya street, Kazan, Russian Federation, 420008\\
$^{\text{\sf 2}}$Steklov Institute of Mathematics at St.~Petersburg, 27 Fontanka, St.~Petersburg, Russian Federation, 191023.}

\corresp{$^\ast$To whom correspondence should be addressed.}

\history{Received on XXXXX; revised on XXXXX; accepted on XXXXX}

\editor{Associate Editor: XXXXXXX}

\abstract{\textbf{Motivation:} Drugs and diseases play a central role in many areas of biomedical research and healthcare. Aggregating knowledge about these entities across a broader range of domains and languages is critical for information extraction (IE) applications. In order to facilitate text mining methods for analysis and comparison of patient's health conditions and adverse drug reactions reported on the Internet with traditional sources such as drug labels, we present a new corpus of Russian language health reviews.\\
\textbf{Results:} The Russian Drug Reaction Corpus (\textsc{RuDReC}) is a new partially annotated corpus of consumer reviews in Russian about pharmaceutical products for the detection of health-related named entities and the effectiveness of pharmaceutical products.
The corpus itself consists of two parts, the raw one and the labelled one. The raw part includes 1.4 million health-related user-generated texts collected from various Internet sources, including social media.
The labelled part contains 500 consumer reviews about drug therapy with drug- and disease-related information. Labels for sentences include health-related issues or their absence. The sentences with one are additionally labelled at the expression level for identification of fine-grained subtypes such as drug classes and drug forms, drug indications, and drug reactions. 
Further, we present a baseline model for named entity recognition (NER) and multi-label sentence classification tasks on this corpus. The macro F1 score of 74.85\% in the NER task was achieved by our RuDR-BERT model. For the sentence classification task, our model achieves the macro F1 score of 68.82\% gaining 7.47\% over the score of BERT model trained on Russian data.\\
\textbf{Availability:} We make the RuDReC corpus and pretrained weights of domain-specific BERT models freely available at \urlstyle{tt}\url{https://github.com/cimm-kzn/RuDReC}\\
\textbf{Contact:} \href{elvtutubalina@kpfu.ru}{elvtutubalina@kpfu.ru}\\
\textbf{Supplementary information:} Supplementary data are available at \textit{Bioinformatics}
online.}

\maketitle

\section{Introduction}
\label{intro}
In this work, we describe the design, composition, and construction of a large dataset of user-generated texts (UGTs) about pharmaceutical products in Russian. Similar to the Food and Drug Administration (FDA) in the U.S. and the Therapeutic Goods Administration (TGA) in Australia, the Federal Service for Surveillance in Healthcare (\emph{Roszdravnadzor}) in Russia accumulates data provided by volunteer reports on the risks of taking various medicines in order to ensure their safe use. Since some particular medications may interact with others in a non-obvious way, creating and using such resources leads to significant difficulties. Information from online sources is considered to be a valuable source for \emph{Roszdravnadzor} or pharmaceutical companies in order to correct the use of a drug when necessary. Thus, our corpus has been designed with the explicit purpose to facilitate the methods for learning complex knowledge of primary interactions between different drugs, diseases, and adverse reactions.

\begin{figure*}[!t]
   \centering
     \includegraphics[width=0.98\linewidth]{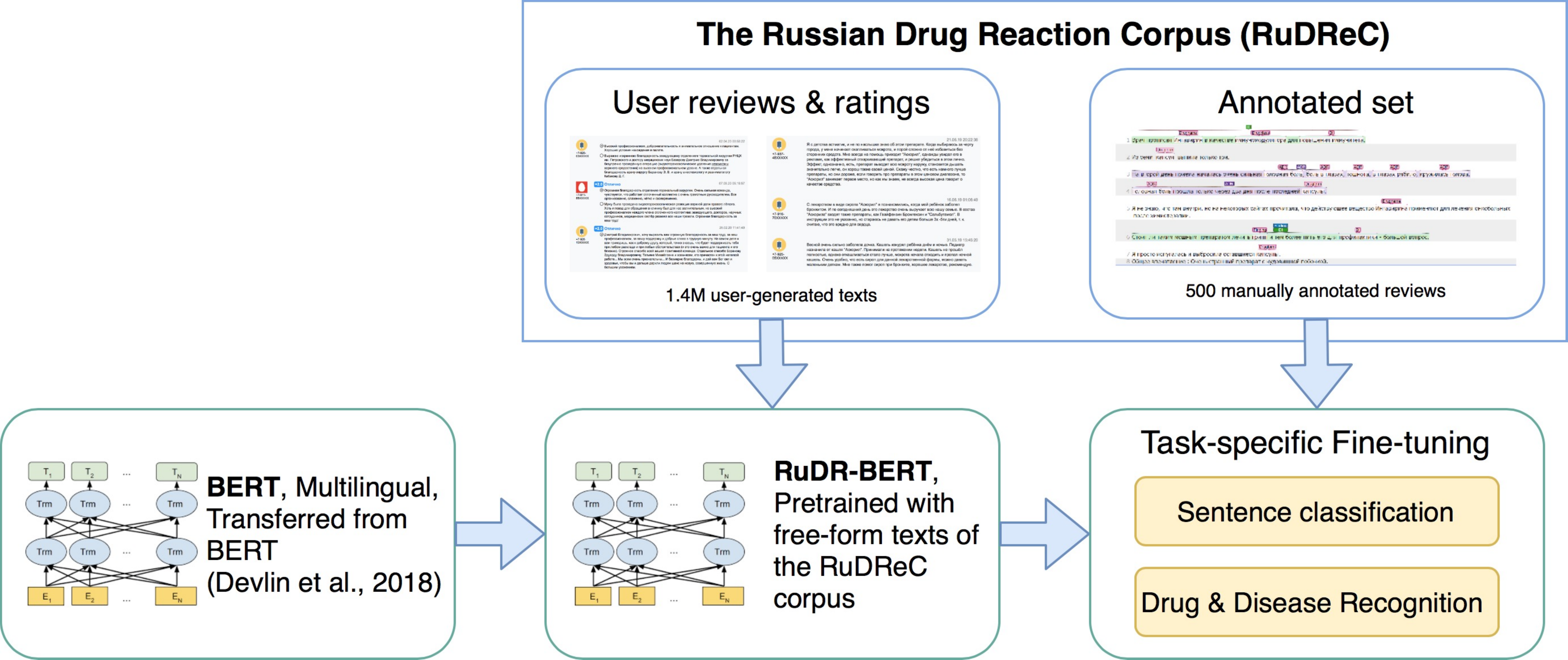}
\caption{Overview of our study: (i) creating the raw and annotated parts of the \emph{RuDReC} corpus, (ii) training a domain-specific version of BERT (RuDR-BERT) on collected texts, and (iii) developing baselines and presenting evaluation results.}\label{fig:annot_example}  
\end{figure*}

Figure~\ref{fig:annot_example} shows a brief overview of our study. The corpus, which we call the \emph{Russian Drug Reaction Corpus} (\textsc{RuDReC}), contains an aggregation of texts of the patients' feedback on the use of drugs in various therapeutic groups or their experience with the healthcare system in general; we have taken care to ensure that we have collected representative samples intended for training advanced machine learning methods. Recent advances in deep contextualized representations via language models such as BERT~\citep{devlin2019bert} or domain-specific biomedical models such as BioBERT~\citep{lee2019biobert} offer new opportunities to improve the models for classification and entity recognition. Our primary goal has been, therefore, to construct a large (partially) annotated corpus in order to stimulate the development of automated text mining methods for finding meaningful information in the patients' narratives in the Russian language.

The \textsc{RuDReC} corpus is meaningfully divided into two parts that are very different in size. The larger part is a raw corpus of 1.4M health comments that can be used to train modern distributed semantics models whose training is based on self-supervised objectives such as the next token prediction (as in, e.g., \emph{word2vec}) or predicting masked tokens (as in, e.g., BERT). The second, smaller part, contains 500 richly annotated reviews to allow the training of downstream task-specific models. The primary downstream tasks in our case are named entity recognition and multi-label classification. The labeling in the second part consists of two main components: sentence labels and entity labels. We have split the review posts into sentences and labeled them for the presence of drug indications and symptoms of a disease (DI), adverse drug reactions (ADR), drug effectiveness (DE), drug ineffectiveness (DIE). In the entity identification phase, we identified and extracted 6 entity types: drug names, drug classes, drug forms, ADR, DI, and Findings. In total, we have labeled 2202 sentences and 4566 entities.

The resulting dataset and pretrained weights of domain-specific BERT have been made freely available for researchers at \urlstyle{tt}\url{https://github.com/cimm-kzn/RuDReC}. We hope that this new resource will intensify research on multilingual IE on adverse drug events and drug effectiveness based on the data from patient narratives. The paper is organized as follows: Section~\ref{sec:related} discusses related work; Section~\ref{sec:main} introduces the \textsc{RuDReC} corpus, describes it qualitatively and quantitatively and shows the details of model training; Section~\ref{sec:eval} presents the results of our evaluation across two downstream tasks (sentence classification and named entity recognition), Section~\ref{sec:limit} shows some limitations of our approach, and Section~\ref{sec:concl} concludes the paper.

\section{Related Work}\label{sec:related}

Many systems for disease and chemical entity recognition from scientific texts have been developed over the past fifteen years. This task is traditionally formulated as a sequence labeling problem and solved with Conditional Random Fields (CRF) that use a wide variety of features: individual words or lemmas, part-of-speech tags, suffixes and prefixes, dictionaries of medical terms, cluster-based and distributed representations, and others~\citep{lee2016audis,gu2016chemical,miftahutdinov2017identifying}. 

In contrast to biomedical literature, research into the processing of user-generated texts (UGTs) about drug therapy has not reached the same level of maturity. Starting from 2014, some studies began to utilize the powers of social media and deep learning (especially suitable for training on large available datasets that are the main advantage of using UGTs) for pharmacovigilance purposes; in particular, researchers have considered the problems of text (post) classification and extraction of adverse drug reactions (ADRs)~\citep{karimi2015cadec,10.1093/jamia/ocy114,alvaro2017twimed,zolnoori2019systematic}. Recent studies primarily employ neural architectures; in particular, \citet{tutubalina2017combination,dang2018d3ner,giorgi2019towards} exploit LSTM-CRF models with domain-specific word embeddings, while~\citet{ecir2020,lee2019biobert} use BERT-based architectures for named entity recognition. 



The CSIRO Adverse Drug Event Corpus (\textsc{CADEC}) dataset collected by  \cite{karimi2015cadec} became a \textit{de facto} standard for the extraction of health-related entities such as ADRs from user reviews. It contains 1253 medical forum posts taken from the \emph{AskaPatient} web portal\footnote{\url{https://www.askapatient.com}} about 12 drugs divided into two categories: \emph{Diclofenac} and \emph{Lipitor}. 
All posts were annotated manually by medical students and computer scientists who labeled five types of entities, including ADRs and names of medicines or drugs. Average inter-annotator agreement rates computed over a subset of 55 user posts with related span matching and tag settings showed that agreement across four annotators in a subset of \emph{Diclofenac} posts was approximately 78\%, while the agreement between two annotators in a subset of \emph{Lipitor} posts was approximately 95\%. 

The Psychiatric Treatment Adverse Reactions (\textsc{PsyTAR}) corpus \citep{zolnoori2019systematic} is also an open source corpus of user-generated posts taken from \emph{AskaPatient}. This dataset includes 887 posts about four psychiatric medications from two classes: (i) \emph{Zoloft} and \emph{Lexapro} from the Selective Serotonin Reuptake Inhibitor (SSRI) class and (ii) \emph{Effexor} and \emph{Cymbalta} from the Serotonin Norepinephrine Reuptake Inhibitor (SNRI) class. In contrast with the CADEC dataset, first, the authors labeled sentences in the posts for the presence of ADRs, withdrawal symptoms (WD), sign/symptoms/illness (SSI), drug indications (DI), drug effectiveness (DE), and drug ineffectiveness (DIE). Second, sentences were annotated with four types of entities: ADR, WD, DI, SSI. Two of the annotators were pharmacy students, and two annotators had a background in health sciences. The resulting pairwise agreement for a strict match was 0.86 for the entire dataset, ranging from 0.81 for the WD class to 0.91 for DI. 

\cite{Shelmanov2015560} created a corpus of clinical notes in the Russian language available for research purposes. The corpus contains 112 fully annotated texts and 45000 tokens from a multi-disciplinary pediatric center. The authors extended an annotation scheme from the CLEF eHealth 2014 Task 2~\citep{suominen2013overview}. Apart from disease mentions, physicians also annotated and verified the mentions of drugs, treatments, and symptoms. A total of 7600 entities of 7 types were identified. The number of entities for each category and inter-annotator agreement rates were not provided, and the annotators did not perform terminology association. The authors developed a knowledge-based method using a set of rules and thesauri, adopting the Russian translation of Medical Subject Headings (MeSH) and the State Register of Drugs (SRD).  

The \emph{Drug Semantics} dataset~\citep{moreno2017drugsemantics} contains 5 summaries of product characteristics in Spanish from an open access repository that belongs to the Spanish Agency for Medicines and Health Products (AEMPS). Each summary concentrates on one of five drugs: \emph{Aspirin}, \emph{Acetaminophen}, \emph{Ibuprofen}, \emph{Atorvastatin}, or \emph{Simvastatin}. The texts were annotated with 10 entity types by a registered nurse and two students pursuing a degree in nursing. The pairwise F-measure between annotators was computed as an agreement measure, and the authors observed the highest agreement for drug-related entities such as Medicament, Excipient, Unit Of Measurement, Drug and Pharmaceutical Form ($F>80\%$), moderate to a substantial agreement for Food, Disease, and Route ($F>60\%$), and weak agreement for Therapeutic Action (F $>$ 10\%). They concluded that agreement rates are comparable with what has been shown for English corpora. 

The \emph{Twitter and PubMed Comparable} corpus (TwiMed) \citep{alvaro2017twimed} is the only open source corpus that contains two sources of information annotated at the entity level by the same experts (pharmacists) using the same set of guidelines. This dataset includes 1000 tweets and 1000 PubMed sentences retrieved using a set of 30 different drugs. This corpus contains annotations for 3144 entities (drugs, symptoms, and diseases), and 5003 attributes of entities (polarity, person, modality, exemplification, duration, severity, status, sentiment). In this case, there was a lower agreement in the annotation of tweets than in the annotation of PubMed sentences, most likely due to the noisy nature of tweets. The annotators did not perform terminology association. We note that the total number of sentences and tweets in the TwiMed corpus is three times smaller than in the CADEC and PsyTAR corpora. 

To sum up, most existing research on information retrieval for drug-related events deals with user reviews, tweets, and clinical records in English~\citep{alvaro2017twimed}; exceptions include studies working with summaries of product characteristics in Spanish \citep{moreno2017drugsemantics}, Russian clinical notes from a multi-disciplinary pediatric center \citep{Shelmanov2015560}, and a French corpus of free-text death certificates~\citep{neveol2017clef, neveol2018clef}. Table~\ref{tab:corpora} presents basic statistics of existing relevant corpora.

There exist very few Russian corpora with annotations of the presence of drug reactions at the level of sentences. \citet{alimova2017machine} proposed a Russian corpus of user reviews from \emph{Otzovik.com} with four types of sentence annotations: indication, beneficial effect, adverse drug reaction, other. Recently, the SMM4H 2020 Task\footnote{\url{https://healthlanguageprocessing.org/smm4h-sharedtask-2020/}} presented a multilingual corpus of tweets (including Russian-language tweets) annotated with the presence of ADRs. To our knowledge, the \textsc{RuDReC} corpus is the first large (partially) annotated corpus of posts about pharmaceutical products in Russian. 

\begin{table*}[!t]
\processtable{Basic statistics of existing drug-related text corpora.\label{tab:corpora}} {\begin{tabular}{@{}p{3.4cm}p{2.cm}p{1.7cm}p{1.7cm}p{7.cm}@{}}
\toprule 
\textbf{Corpus} & \textbf{Text type} & \textbf{No. of texts} & \textbf{No. of entities} & \textbf{Annotations} \\ 
\midrule
\multicolumn{5}{c}{\textbf{English language corpora}} \\ \hline
CADEC~\citet{karimi2015cadec} & User reviews & 1253 & 9111 & ADR, Disease, Symptom, Finding, Drug\\ 
PsyTAR~\cite{zolnoori2019systematic} & User reviews  & 887 & 7414 & ADR, Withdrawal symptom, symptoms, drug indication (DI) \\ 
TwiMed~\cite{alvaro2017twimed} & \raggedright Sentences from abstracts, tweets & 2000 & 3144 &  Drug, Symptom, Diseases\\ 
\hline
\multicolumn{5}{c}{\textbf{Spanish language corpora}} \\ \hline
Drug Semantics~\cite{moreno2017drugsemantics} & \raggedright Summaries of Product Characteristics & 5 & 2241 & Disease, Drug, Unit of Measurement, Excipient, Chemical Composition, Pharmaceutical Form, Route, Medicament Food and Therapeutic Action \\ 
\hline
\multicolumn{5}{c}{\textbf{Russian language corpora}} \\ \hline
\cite{Shelmanov2015560} & Clinical notes  & 112 & 7600 & Disease, Symptom, Drug, Treatment, Body location, Severity, Course \\ 
\cite{alimova2017machine} & Reviews' sentences & 370 & N/A & Sentences from reviews annotated with the presence of Indication, Beneficial effect, ADR, Other \\ 
SMM4H 2020 Task & Tweets & 9515 & N/A & Tweets annotated with presence of ADRs \\ 
\textsc{RuDReC} (ours) & User reviews & 500 & 4566 & DI, ADR, Finding, Drug name, Drug class, Drug form \\
\botrule
\end{tabular}}{}
\end{table*}

\section{The \textsc{RuDReC} corpus}\label{sec:main}

Our goal in this work is three-fold: 
\begin{enumerate}[(1)]
    \item~create an open access corpus, which we call \textsc{RuDReC} that would conform to annotation guidelines based on the annotators' insights and existing English corpora such as CADEC and PsyTAR;
    \item~collect a large dataset of free-form health-related UGTs in order to ensure diversity of drug classes that are defined by their therapeutic use;
    \item~develop a domain-specific language representation model, pretrained on the raw texts from the collected corpus and baselines for sentence classification and entity recognition tasks.
\end{enumerate}
Our manually annotated corpus contains 5 sentence labels and 6 different entity types as shown in Tables \ref{tab:sentlabels} and \ref{tab:entities}, respectively. 

\begin{table}[!t]
\processtable{Definitions for sentence labels annotated in the patients' comments\label{tab:sentlabels}} {\begin{tabular}{@{}p{1.9cm}p{6cm}@{}}
\toprule 
\textbf{Sentence label} & \textbf{Definition} \\
\midrule
\textsc{DE} & A sentence is labeled as Drug Effectiveness (DE) if it contains an explicit report about treated symptoms or that the patient's condition has improved after drug use.\\
\textsc{DIE} & A sentence is labeled as Drug Ineffectiveness (DIE) if it contains a direct report that the patient’s health status became worse or did not change after the drug usage. \\
\textsc{DI} & A sentence is labeled as Drug Indication (DI) if it contains any indication/symptom that specifies the reason for taking/prescribing the drug.\\
\textsc{ADR} & A sentence is labeled as Adverse Drug Reaction (ADR) if it contains mentions of undesirable, untoward medical events that occur as a consequence of drug intake.\\
\textsc{Finding} & A sentence is labeled as Finding if it describes disease-related events that are not experienced or denied by the reporting patient or his/her family members. These sentences often describe a patient's medical history, drug label, or absence of expected drug reactions. \\
\botrule
\end{tabular}}{}
\end{table}

\begin{table}[!t]
\processtable{Definitions for entity types identified in patient comments\label{tab:entities}} {\begin{tabular}{@{}p{1.9cm}p{6cm}@{}}
\toprule 
\textbf{Entity type} & \textbf{Definition} \\
\midrule
\textsc{Drugname} & Mentions of the brand name of a drug or product ingredients/active compounds.  \\
\textsc{Drugclass} & Mentions of drug classes such as \textit{anti-inflammatory} or \textit{cardiovascular}.  \\
\textsc{Drugform} & Mentions of routes of administration such as \textit{tablet} or \textit{liquid} that describe the physical form in which medication will be delivered into patient's organism.  \\
\textsc{DI} & Any indication/symptom that specifies the reason for taking/prescribing the drug. \\
\textsc{ADR} & Mentions of untoward medical events that occur as a consequence of drug intake and are not associated with treated symptoms. \\
\textsc{Finding} & Any DI or ADR that was not directly experienced by the reporting patient or his/her family members, or related to medical history/drug label, or any disease entities if the annotator is not clear about type.\\
\botrule
\end{tabular}}{}
\end{table}

\begin{figure*}[!t]
  \centering
     \includegraphics[width=0.98\linewidth]{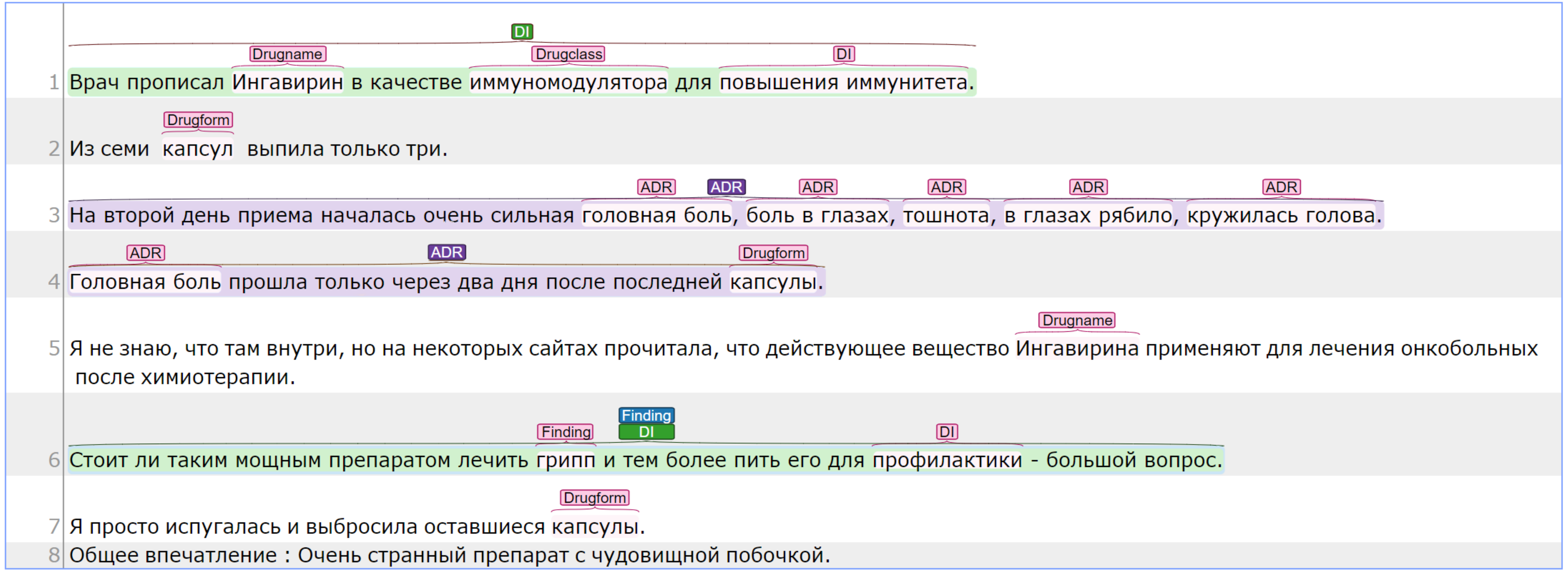}
\caption{Example of sentence and entity annotation.}\label{fig:INCEpTION}  
\end{figure*}

Figure~\ref{fig:INCEpTION} shows sample annotations produced using INCEpTION as the annotation platform~\citep{tubiblio106270}. It is important to note that we have obtained all reviews without accessing password-protected information; all data from our corpus is publicly available on the Internet.  

\subsection{Annotation}

\subsubsection{Data source} 
For the annotation process, we have utilized user posts in Russian from a popular and publicly accessible source \emph{Otzovik.com}, which collects the patients' self-reported experiences for a wide range of medications. Each user fills out a form containing the drug description (including the reason for taking it), drug class, year of purchase, its route of administration, perceived efficiency, and side effects, and information about the disease. Users are also asked to rate the overall drug satisfaction from 1 (low) to 5 (high). The reviews are written in Russian; as is usually the case with user-generated texts, they do not necessarily have perfect grammar and may contain informal language patterns specific for different regions of Russia and other Russian-speaking countries. 

\subsubsection{Annotation Guidelines}
Our annotation process consisted of two stages. At the first stage, annotators with a background in pharmaceutical sciences were asked to read 400 reviews and highlight all spans of text, including drug names and patient's health conditions experienced before, during, or after the drug use. The objective of the first stage of the annotation process was to perform preliminary annotation across a set of reviews in order to choose the best annotation scheme. The authors informed the annotators  with an analysis of existing annotation schemes for English language corpora~\citep{karimi2015cadec,alvaro2017twimed}. At the second stage, annotators were asked to screen existing annotations and annotate new texts on an extended set of reviews. 

At the first stage, the process of identification and extraction of entities' spans was conducted by four annotators with a background in pharmaceutical sciences from the I.M. Sechenov First Moscow State Medical University. Our analysis of existing corpora shows two main types of entities common to all schemes: \textsc{Drug} and \textsc{Disease}. After several discussions, annotators defined the following \textsc{Disease} subtypes:
\begin{inparaenum}[(1)]
\item disease name;
\item indication (Indication);
\item positive dynamics after or during taking the drug (BNE-Pos);
\item negative dynamics after the start or some period of using the drug (ADE-Neg);
\item the drug does not work after taking the course (NegatedADE);
\item deterioration after taking a course of the drug (Worse).
\end{inparaenum}
As \textsc{Drug} subtypes, annotators have chosen:
\begin{inparaenum}[(1)]
\item drug names,
\item drug classes, and
\item drug forms.
\end{inparaenum}

The posts were divided between the annotators, and 100 documents and annotation guidelines were given to another annotator from the department of pharmacology of the Kazan Federal University for the purpose of calculating the inter-annotator agreement. We note that this annotator did not interact with other annotators in discussions about the annotation scheme. Two metrics were used in our calculation of relaxed agreement for \textit{Disease} and \textit{Drug} entities, as described in~\citep{karimi2015cadec}. When annotation and span settings were both relaxed, the average agreement was approximately 70\%.

After completing the annotation process at the first stage, three of the authors screened the annotations. We came to several conclusions based on the results. First, there were relatively few examples of Worse and ADE-Neg types (198 examples in total). Second, entities of ineffective type were longer in comparison with other entity types: the average length of ineffective type entities was 15 words, while, e.g., ADRs had an average of 5 words. Finally, the BNE-Pos entity types contained a lot of overly broad entities that were not related to medical concepts, such as ``helped'',  ``effective'', and so on.

To mitigate these problems, we made several changes to the annotation scheme. First, we combined \textit{Worse} and \textit{ADE-Neg} with \textit{NegatedADE} entity types into a single class \textit{Drug Ineffectiveness} (\textsc{DIE}) and spanned \textsc{DIE} annotation on the sentence level, similar to the PsyTAR corpus. Second, we spanned \textit{BNE-Pos} entities on the sentence level and renamed them to \textit{Drug Effectiveness} (\textsc{DE}), also in agreement with the PsyTAR corpus. Finally, following the CADEC corpus, we combined the \emph{Indication} and \emph{Disease} entity types into a single \textit{Drug Indication} (\textsc{DI}) type.

At the second stage, two annotators from the Kazan Federal University were asked to continue the annotation process according to sentence classes and entity types presented in Tables~\ref{tab:sentlabels} and \ref{tab:entities}. After completing the annotation process, two of the authors screened the annotations to correct span mistakes.

\subsection{Analysis of the Annotated Set}

Our dataset includes reviews about four groups of drugs:
\begin{enumerate}[(1)]
\item~sedatives (brain and nervous system);
\item~nootropics (brain and nervous system);
\item~immunomodulators (immune disease);
\item~antivirals (infections).
\end{enumerate}
Sedatives and nootropics both belong to the neurotropic group of drugs, i.e., drugs that have an effect on the central and peripheral nervous systems. This group includes antidepressants, mood stabilizers, nootropics, and sedatives. Immunomodulators, in particular immunostimulants and immunosuppressants, are substances that modify the immune response and affect immunocompetent cells. Antiviral drugs are intended for the treatment of various viral diseases (influenza, herpes, HIV infection, etc.); they are also used for preventive purposes.

The annotated corpus consists of 500 reviews about drugs from these four groups. Reviews were selected randomly for annotation. The examples of annotated entities for each group are presented in Supplementary Table S1. 

\begin{figure}[!tpb]
\centerline{\includegraphics[width=0.6\linewidth]{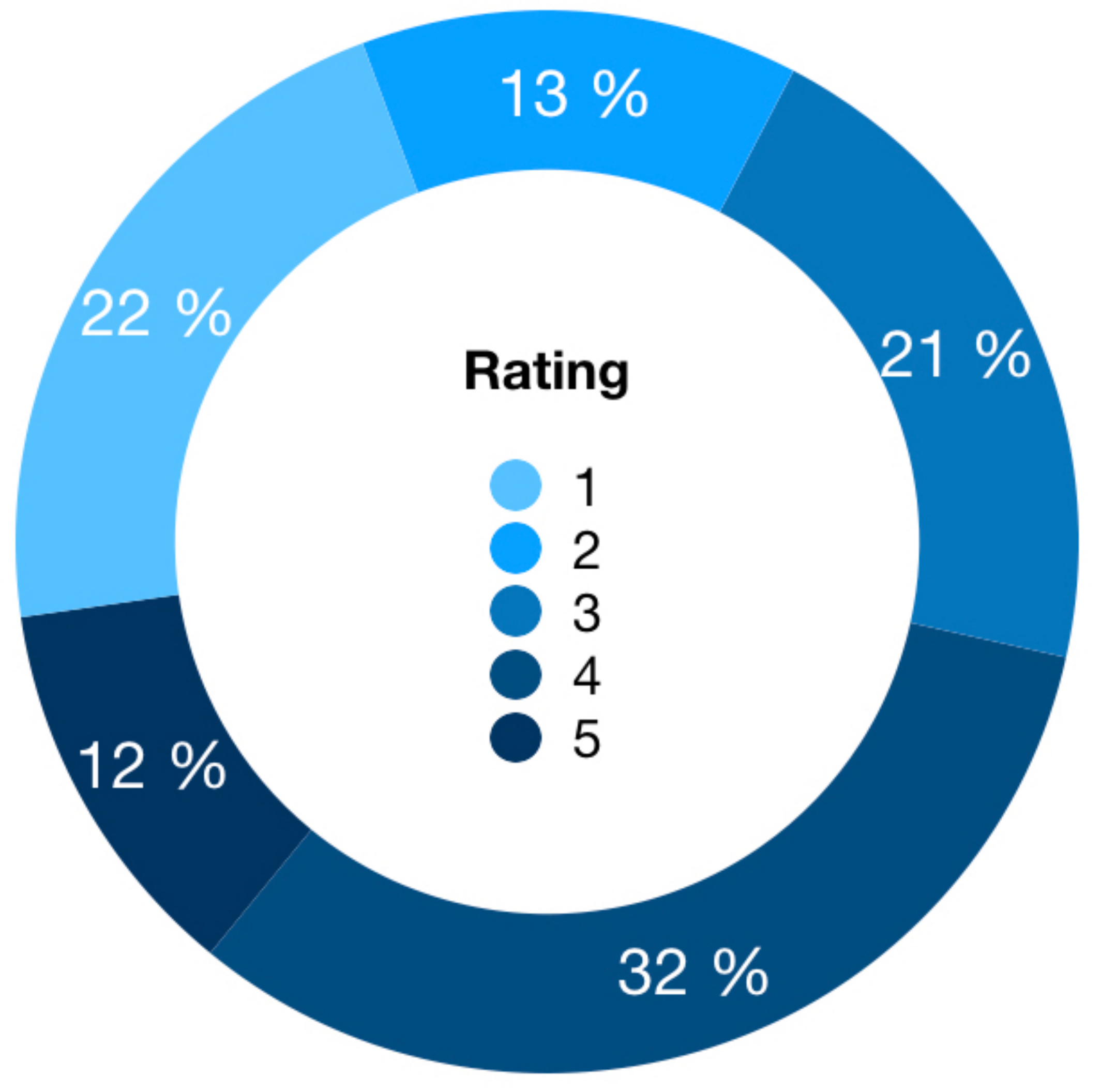}}
\caption{Statistics on ratings in the annotated corpus (filled by reviews' authors).}\label{fig:pharm_rating}
\end{figure}

Figures \ref{fig:pharm_group} and \ref{fig:pharm_rating} present statistics on therapeutic groups and ratings in our corpus, respectively. Every user fills out this information when writing a review. The majority of the reviews are describing the antiviral drugs, which are of the most common ones used in everyday life. The second by number group is sedatives and antidepressants, which are on the raise in recent years.

Another interesting feature of the presented statistics is that the prevalence of the highest rating (5) is not overpowering the other ratings, which are more or less uniformly distributed. This is a common feature that the intermediate rating is mostly skipped in many domains, but the collected data is showing unusual uniformity.  

\begin{figure}[!tpb]
\centerline{\includegraphics[width=0.6\linewidth]{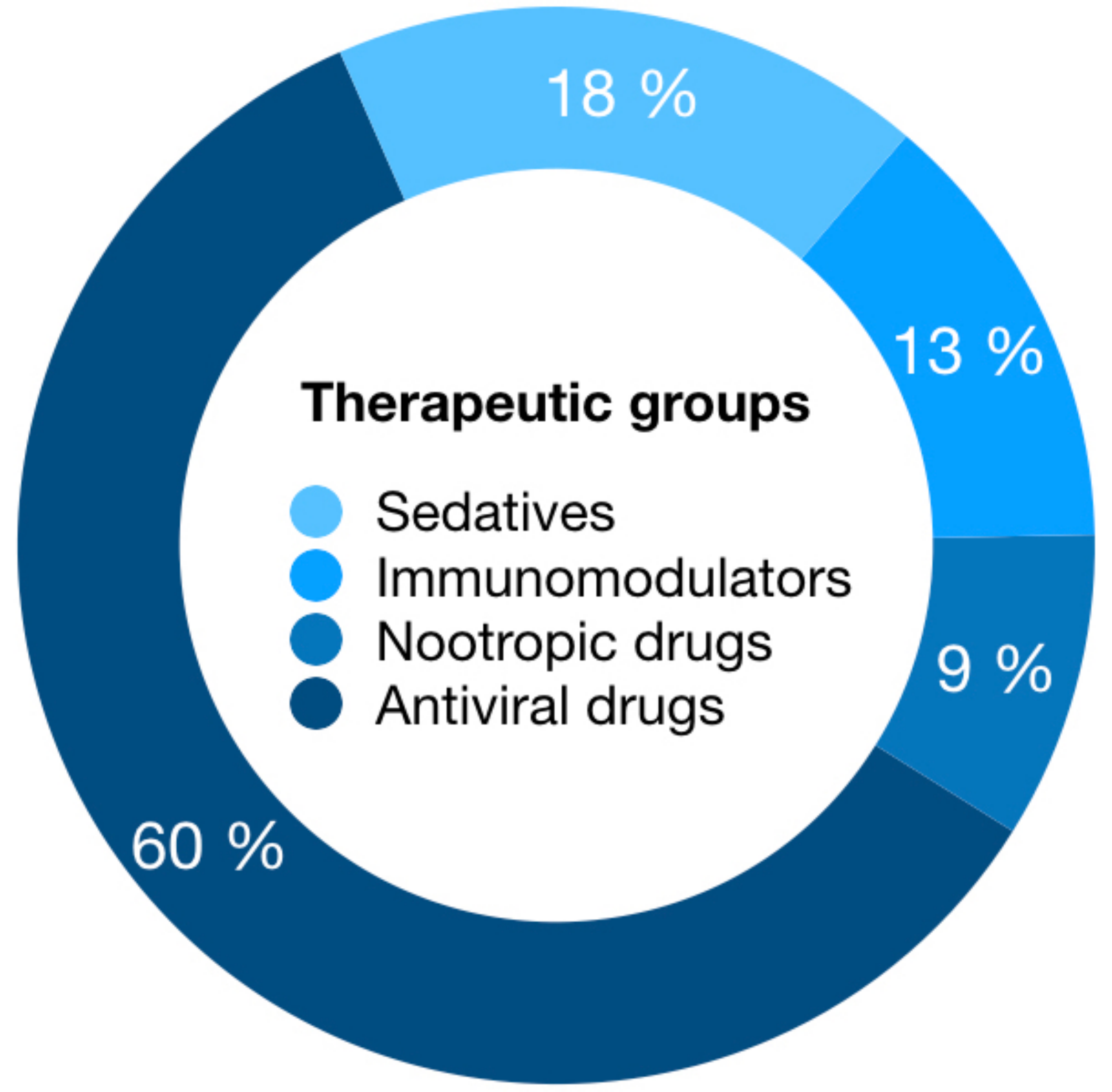}}
\caption{Statistics on therapeutic groups in the annotated corpus (filled by reviews' authors).}\label{fig:pharm_group}
\end{figure}

Table \ref{tab:stats_texts} lists the statistics for the annotated corpus part as a whole, as well as one for each group of drugs. There are several interesting features one could note here. First of all, immunomodulatory drugs have longer reviews in terms of both the sentences and tokens. The average length is 30\% larger than for any other group, and the maximal length is up to twice larger, although the minimal length is the same as for other groups. Second, the average number of sentences in Russian reviews is higher than in the English CADEC and PsyTAR corpora ($9.71$ vs. $6$).

\begin{table}[!th]
\processtable{Basic statistics on reviews, sentences, and tokens\label{tab:stats_texts}} {\begin{tabular}{@{}p{2.5cm}p{.9cm}p{.8cm}p{1.2cm}p{.7cm}p{.7cm}@{}}
\toprule 
& \bf Entire Corpus & \bf Seda-tives & \bf Immuno-modulators & \bf Noo-tropics & \bf Anti-virals \\
\midrule
No. of reviews & 500 & 90 & 67 & 46 & 297 \\ 
Total no. of sentences & 4855 & 829 & 813 & 410 & 2803 \\ 
Avg no. of sentences in each review & 9.71 & 9.21 & 12.13 & 8.91 & 9.44\\ 
No. of sentences in each review (range) & 1-35 & 2-22 & 2-35 & 3-17 & 1-25\\ 
Total no. of tokens & 68036 & 11536 & 12217 & 5930 & 38353\\ 
Avg no. of tokens (words) in each review & 136.07 & 128.17 & 182.34 &  128.91 & 129.13\\ 
\botrule
\end{tabular}}{}
\end{table}

Table~\ref{tab:stats_sentences} presents the frequency of annotated sentences in the entire corpus as well as in each drug group. There are several features that should be mentioned regarding these annotations. There are interesting disproportionalities in the frequencies (normalized columns) of different types of labels. The immunomodulators group has the lowest representation of adverse drug reactions, while the antidepressants (sleeping) the highest one. 

\begin{table*}[!t]
\processtable{Number of sentences annotated in the entire corpus and each therapeutic group\label{tab:stats_sentences}}
{\begin{tabular}{@{}l|cc|cc|cc|cc|cc@{}}\toprule 
& \multicolumn{2}{p{2.5cm}|}{\bf Entire Corpus} & \multicolumn{2}{p{1.5cm}|}{\bf Sedatives} & \multicolumn{2}{p{2.5cm}|}{\bf Immunomodulators} & \multicolumn{2}{p{1.5cm}|}{\bf Nootropics} & \multicolumn{2}{p{1.5cm}}{\bf Antivirals} \\
& Raw & Norm. & Raw & Norm. & Raw & Norm. & Raw & Norm. & Raw & Norm. \\
\midrule
\textsc{DI} & 949 &1.90 & 182& 2.02& 132&1.97 & 83&1.80 & 552&1.86 \\ 
\textsc{ADR} & 379&0.78 & 100	&1.11& 27 &0.40& 42 &0.91& 210&0.71\\ 
\textsc{FINDING} & 172&0.34 & 36&0.40 & 25&0.37 & 20&0.43 & 91&0.31\\ 
\textsc{DE} & 424&0.85 & 86&0.96 & 69&1.03 & 53&1.15 & 216&0.73 \\ 
\textsc{DIE} & 278&0.56 & 45& 0.50& 35& 0.52  & 26&0.57 & 172&0.58 \\ 
\hline
All &  2202 &4.40&  449 &4.99&  288 &4.30& 224 &4.87&  1241 & 4.18 \\
\botrule
\end{tabular}}{}
\end{table*}

Table \ref{tab:stats_entities} presents the statistics of annotated entities in the entire corpus as well as in each drug group. The drug class and drug form labels are surprisingly scarce in the nootropic group. The most common among others DI class is in the antidepressant group.

\begin{table*}[!t]
\processtable{Number of entities annotated in the entire corpus and each therapeutic group\label{tab:stats_entities}}
{\begin{tabular}{@{}l|cc|cc|cc|cc|cc@{}}
\toprule 
& \multicolumn{2}{p{2.5cm}|}{\bf Entire Corpus} & \multicolumn{2}{p{1.5cm}|}{\bf Sedatives} & \multicolumn{2}{p{2.5cm}|}{\bf Immunomodulators} & \multicolumn{2}{p{1.5cm}|}{\bf Nootropics} & \multicolumn{2}{p{1.5cm}}{\bf Antivirals} \\
& Raw & Norm. & Raw & Norm. & Raw & Norm. & Raw & Norm. & Raw & Norm. \\
\midrule
\textsc{DRUGNAME} & 1043& 2.07& 200&2.22& 151&2.25 & 95&2.07 & 597&2.01\\ 
\textsc{DRUGCLASS} & 330 &0.66& 79 &0.88& 64&0.96 & 8&0.17 & 179&0.60\\ 
\textsc{DRUGFORM} & 836 &1.67& 155 &1.72& 163 &2.43& 35 &0.76& 483 &1.63\\ 
\textsc{DI} & 1401 &2.80& 293 &3.26& 191 &2.85& 116 &2.52& 801 &2.70\\ 
\textsc{ADR} & 720 &1.44& 202 &2.24& 43 &0.64& 93 &2.02& 382 &1.29\\ 
\textsc{FINDING} & 236 &0.47& 50 &0.56& 33 &0.49& 26 &0.54& 127 &0.43\\ 
\hline
All & 4566 &9.13& 979 &10.88& 645 &9.62& 372 &8.09& 2570 &8.65\\ 
\botrule
\end{tabular}}{}
\end{table*}

\subsection{A large collection of Health Reviews}

Text collections used for training domain-specific BERT were obtained by web page crawling. User reviews were collected from popular medical web portals. 
These online resources mostly contain drug reviews about pharmaceutical products, health facilities, and pharmacies. Duplicate comments were removed. The statistics on this part of the \textsc{RuDReC} corpus are given in Table~\ref{tab:raw-stats}. The collection contains 1.4 million of patient narrative texts, 1{,}104{,}054 unique tokens, and 193{,}529{,}197 tokens in total.

\begin{table}[!t]
\processtable{Text collection statistics for web-based comments\label{tab:raw-stats}} {\begin{tabular}{@{}lcr@{}}
\toprule 
\textbf{Category for reviewing} & \textbf{Written by} & \textbf{Number of texts} \\ 
\midrule
Pharmaceutical products & users & 261{,}983\\ 
Beauty products & users & 466{,}199 \\ 
Drugs & doctors & 7{,}451 \\ 
Drugs & users & 31{,}500 \\ 
Health facilities \& pharmacies & users & 642{,}178\\ 
\hline
\textbf{Total} & & \textbf{1{,}409{,}311} \\
\botrule
\end{tabular}}{}
\end{table}

\subsection{Pre-training \& Fine-tuning domain-specific BERT}

We used the multilingual version of BERT-base (Multi-BERT) as initialization for training domain-specific BERT further called \textbf{RuDR-BERT}. 

Similar to~\citep{lee2019biobert}, we observed that 800K and 840K pretraining steps were sufficient. This roughly corresponds to a single epoch on each corpus. The batch size was set to 32 examples. Other hyperparameters such as learning rate scheduling for pretraining RuDR-BERT are the same as those for Multi-BERT unless stated otherwise. We decided to adopt the initial vocabulary of Multi-BERT for preprocessing in both pretraining corpora and fine-tuning sets. The language model was fine-tuned using a BERT implementation from \urlstyle{tt}\url{https://github.com/google-research/bert}. We trained \textbf{RuDR-BERT} on a single machine with 8 NVIDIA P40 GPUs. The training of all models took approximately 8 days.

We fine-tuned several BERT models, including RuDR-BERT, on two tasks:
\begin{enumerate}[(i)]
\item~named entity recognition (with entity types as shown in Table~\ref{tab:stats_entities});
\item~sentence classification (the classes are presented in Table~\ref{tab:stats_sentences}).
\end{enumerate}
Following our previous work on NER~\citep{ecir2020}, we utilize different BERT models with a softmax layer over all possible tags as the output for NER. Word labels are encoded with the BIO tag scheme. We note that the model was trained on the sentence level. All NER models were trained without an explicit selection of parameters on the \textsc{RuDReC} corpus. The loss function became stable (without significant decreases) after 35-40 epochs. We use Adam optimizer with polynomial decay to update the learning rate on each epoch with warm-up steps in the beginning. 
For sentence classification, we utilize the Tensorflow implementation of BERT with sigmoid activation over dense output layer and cross-entropy loss function. For each label, we used the sigmoid value of 0.5 as a classification threshold. We fine-tuned each model for 10 epochs with a batch size of 16. We defined the first 10\% of the training steps as warm-up steps.

\section{Experiments and Evaluation}\label{sec:eval}

For our experiments, we used three versions of BERT:
\begin{enumerate}[(1)]
    \item~BERT$_{base}$, the Multilingual Cased (Multi-BERT) pretrained on 104 languages; it has 12 heads, 12 layers, 768 hidden units per layer, and a total of 110M parameters;
    \item~RuBERT, the Russian Cased BERT pretrained on the Russian part of Wikipedia and news data; it has 12 heads, 12 layers, 768 hidden units per layer, and a total of 180M parameters; Multi-BERT was used for initialization, while the vocabulary of Russian subtokens was built on the training dataset~\citep{kuratov2019adaptation};
    \item RuDR-BERT, Multilingual Cased BERT pretrained on the raw part of the \textsc{RuDReC} corpus (1.5M reviews); Multi-BERT was used for initialization, and the vocabulary of Russian subtokens and parameters are the same as in Multi-BERT.
\end{enumerate}


\subsection{Multi-label Sentence Classification}

\begin{table*}[!t]
\processtable{Performance of fine-tuned RuDR-BERT on sentence classification with comparison to multi-BERT and RuBERT, measured by F1-score.\label{tab:sent-classif}} {\begin{tabular}{@{}lccccc|c@{}}
\toprule
\textbf{Model} & \textbf{DE} & \textbf{DIE} & \textbf{ADR} & \textbf{DI} & \textbf{Finding} & \textbf{Macro F1-score}  \\
\midrule
\textbf{RuBERT} & 67.7 $\pm$ 2.82 & 62.27 $\pm$ 3.47 & 66.65 $\pm$ 2.96 & 81.63 $\pm$ 2.38 & 28.51 $\pm$ 4.8 & 61.35 $\pm$ 3.28\\
\textbf{Multi-BERT} & 63.61 $\pm$ 4.22 & 60.19 $\pm$ 3.52 & 63.45 $\pm$ 2.61 & 79.58 $\pm$ 4.1 & 24.32 $\pm$ 2.85 & 58.23 $\pm$ 3.46\\
\textbf{RuDR-BERT} & 76.61 $\pm$ 4.08 & 72.06 $\pm$ 5.29 & 74.15 $\pm$ 5.01 & 85.06 $\pm$ 2.49 & 36.24 $\pm$ 6.91 & 68.82 $\pm$ 4.76 \\
\botrule
\end{tabular}}{}
\end{table*}

We compare all models on 5-fold cross-validation in terms of F1-score. The fine-tuning of each model took approximately 1 hour on one NVIDIA GTX 1080 Ti GPU.

Table~\ref{tab:sent-classif} performs the result of RuBERT, Multi-BERT, and fine-tuned RuDR-BERT models in terms of F1-score. According to the results, the following conclusions can be drawn. First, the RuDR-BERT model achieved the best results among other comparable models. Second, the RuBERT model outperformed the Multi-BERT model on 3.12\% in terms of the macro F1-score. The highest improvement was achieved for DE (+4.09\%) and Finding entity types (+4.19\%). Third, the performance of RuDR-BERT on Finding (36.24\%) is significantly lower than on ADR (74.15\%) and DI (85.06\%). It could be explained by similar contexts and a much lower number of training examples.

\subsection{Drug and Disease Recognition}
We compare all models on 5-fold cross-validation in terms of F1-scores computed by exactly matching criteria via a CoNLL script. We trained each model on a single machine with 8 NVIDIA P40 GPUs. The training of all models took approximately 10 hours.

Table~\ref{tab:NER} shows the performance of RuBERT, Multi-BERT, and fine-tuned RuDR-BERT in terms of the F1-score. There are several conclusions to be drawn based on the results in these tables. First, on all types of entities, the domain-specific RuDR-BERT achieves better scores than both RuBERT and Multi-BERT. Second, RuBERT, with a vocabulary of Russian subtokens generated on Wikipedia and news, outperforms Multi-BERT. Third, similarly to sentence classification, the performance of RuDR-BERT on \textsc{Finding} is significantly lower than on \textsc{ADR} and \textsc{DI}. Finally, all models achieve much higher performance for the detection of drugs rather than diseases; it can be explained by boundary problems in multi-word expressions. The average number of tokens on drug-related entities is 1.06, while the average number of tokens on disease-related entities is 1.77. To obtain metrics for disease-related entities, we replaced ADR, DI, and Finding entity types with \textit{Disease} entity type in the gold standard and predicted data. The same procedure was done for drug-related entities except that Drugname, Drugform, and Drugclass were replaced by \textit{Drug}. The RuDR-BERT model achieves the F1-score of 81.34\% on disease-related entities and F1-score of 94.65\% of drug-related entities.  

\begin{table*}[!t]
\processtable{Performance of fine-tuned RuDR-BERT on the recognition of 6 entity types in comparison with Multi-BERT and RuBERT, measured by F1-score with exact matching criteria\label{tab:NER}} {\begin{tabular}{@{}lcccccc|c@{}}
\toprule 
\textbf{Model} & \textbf{ADR} & \textbf{DI} & \textbf{Finding}  & \textbf{Drugclass} & \textbf{Drugform} & \textbf{Drugname} &  \textbf{Macro F1-score} \\ 
\midrule
\textbf{RuBERT} & 54.51 $\pm$ 3.9 & 69.43 $\pm$ 4.98 & 27.87 $\pm$ 5.92  & 92.78 $\pm$ 1.14 & 95.72 $\pm$ 1.38 & 92.11 $\pm$ 1.56 & 72.07 $\pm$ 2.03 \\
\textbf{Multi-BERT} & 54.65 $\pm$ 2.38 & 67.63 $\pm$ 3.62 & 25.75 $\pm$ 7.86 & 92.36 $\pm$ 2.72 & 94.89 $\pm$ 0.97 & 91.05 $\pm$ 0.61 & 71.06 $\pm$ 2.46 \\ 
\textbf{RuDR-BERT} & 60.36 $\pm$ 2.13 & 72.33 $\pm$ 2.12 & 33.31 $\pm$ 7.55  & 94.12 $\pm$ 2.31 & 95.89 $\pm$ 1.82 & 93.08 $\pm$ 1.08 & 74.85 $\pm$ 2.09 \\  
\botrule
\end{tabular}}{}
\end{table*}


\section{Limitations}\label{sec:limit}

There are several issues that may potentially limit the applicability of \textsc{RuDReC}; they are mostly shared with other available datasets.

\paragraph{Validation of drugs by the State Register of Medicines.} 
We believe that automatic systems for extracting meaningful information concerning pharmaceutical products should validate whether the pharmaceutical products have registered with the State Register of Medicines\footnote{\url{https://grls.rosminzdrav.ru/}}. The State Register of Medicines is a list of domestic and foreign medicines, medical prophylactic and diagnostic products registered by the Ministry of Health of Russia. Our annotator from the Department of Pharmacology of Kazan Federal University conducted a manual study of 649 unique product names that review authors put as review titles in their free-form reviews, checking whether the drugs were present in the State Register of Medicines for each product name. The results of this labeling showed that 373 (57.5\%) of the names do have a match in the system and belong to one of the groups from the Anatomical Therapeutic Chemical (ATC) Classification System (J0, D0, G0, A0). Note that this is a preliminary result, and it has not been validated with multiple annotators; however, it indicates the need for an additional validation step for automatic systems.

\paragraph{Normalization challenge.} There are three major international terminologies for the Russian language: Medical Dictionary for Regulatory Activities (MedDRA), Medical Subject Headings (MeSH) thesaurus, and Classification of Diseases (ICD). One challenge is that layperson expressions of disease-related words are fuzzier and broader than the corresponding MedDRA terms. Another challenge is that social media patients discuss different concepts of illness and a wide diversity of drug reactions. Moreover, social network data usually contains a lot of noise, such as misspelled words, incorrect grammar, hashtags, abbreviations, and different variations of the same word. In our dataset, there is no mapping of entity mentions to formal medical terminology, which we leave as future work.

\paragraph{The Risk of fake reports on the Internet.} A recent study by~\citet{smith2018methods} demonstrates that it is possible to harvest and compare ADRs found in social media with those from traditional sources. One major challenge for automatic methods is fact-checking. A similar research question is currently being investigated in the \emph{CLEF-2020 CheckThat! Shared Task 1} that deals with whether a given tweet is trustworthy, i.e., whether it is supported by factual information (the task uses a sample of tweets about COVID-19).

\paragraph{Robustness of trained models.} Our annotated corpus for training NER and classification models includes reviews on several therapeutic groups, but it may not be representative of drugs from other classes, for example, antineoplastic agents. On the other hand, the \textsc{RuDReC} corpus includes a large collection of 1.4M user-generated health reviews about a large assortment of pharmaceutical products and patient experience with hospital care that could improve the robustness of language models. 

\section{Conclusion}\label{sec:concl}

In this work, we present a new open access corpus named \textsc{RuDReC} (Russian Drug Reaction Corpus) for researchers of biomedical natural language processing and pharmacovigilance. In the paper, we have discussed the challenges of annotating health-related Russian comments and have presented several baselines for the classification and extraction of health entities. The \textsc{RuDReC} corpus provides opportunities for researchers in a number of areas in order to:
\begin{enumerate}[(1)]
\item~develop and evaluate text mining models for gathering of meaningful information about drug effectiveness and adverse drug reactions from layperson reports;
\item~analyze and compare variations of reported patient health conditions and drug reactions of different therapeutic groups of medications with drug labels.
\end{enumerate}

We foresee three directions for future work. First, transfer learning and multi-task strategies on several tasks on English and Russian texts remain to be explored. Second, a promising research direction is to try pretraining domain-specific BERT-based models with a custom vocabulary. Third, future research will focus on the creation of mapping between entity mentions and existing multilingual terminologies such as MedDRA and MeSH.

\section*{Acknowledgements}
We thank all annotators for their contribution. We also thank Timur Madzhidov and Valery Solovyev for valuable feedback.

\section*{Funding}
This work has been supported by the Russian Science Foundation grant \# 18-11-00284.


\begin{thebibliography}{}

\bibitem[Alimova {\em et~al.}(2017)Alimova, Tutubalina, Alferova, and
  Gafiyatullina]{alimova2017machine}
Alimova, I., Tutubalina, E., Alferova, J., and Gafiyatullina, G. (2017).
\newblock A machine learning approach to classification of drug reviews in
  russian.
\newblock In {\em 2017 Ivannikov ISPRAS Open Conference (ISPRAS)\/}, pages
  64--69. IEEE.

\bibitem[Alvaro {\em et~al.}(2017)Alvaro, Miyao, and Collier]{alvaro2017twimed}
Alvaro, N., Miyao, Y., and Collier, N. (2017).
\newblock Twimed: Twitter and pubmed comparable corpus of drugs, diseases,
  symptoms, and their relations.
\newblock {\em JMIR public health and surveillance\/}, {\bf 3}(2).

\bibitem[Dang {\em et~al.}(2018)Dang, Le, Nguyen, and Vu]{dang2018d3ner}
Dang, T.~H., Le, H.-Q., Nguyen, T.~M., and Vu, S.~T. (2018).
\newblock D3ner: biomedical named entity recognition using crf-bilstm improved
  with fine-tuned embeddings of various linguistic information.
\newblock {\em Bioinformatics\/}, {\bf 34}(20), 3539--3546.

\bibitem[Devlin {\em et~al.}(2019)Devlin, Chang, Lee, and
  Toutanova]{devlin2019bert}
Devlin, J., Chang, M.-W., Lee, K., and Toutanova, K. (2019).
\newblock Bert: Pre-training of deep bidirectional transformers for language
  understanding.
\newblock In {\em Proceedings of the 2019 Conference of the North American
  Chapter of the Association for Computational Linguistics: Human Language
  Technologies, Volume 1 (Long and Short Papers)\/}, pages 4171--4186.

\bibitem[Giorgi and Bader(2019)Giorgi and Bader]{giorgi2019towards}
Giorgi, J. and Bader, G. (2019).
\newblock Towards reliable named entity recognition in the biomedical domain.
\newblock {\em Bioinformatics (Oxford, England)\/}.

\bibitem[Gonzalez-Hernandez {\em et~al.}(2018)Gonzalez-Hernandez, Sarker,
  Nenadic, Belousov, Friedrichs, Ginter, Mehryary, Hakala, de~Bruijn, Mohammad,
  Kiritchenko, Rios, Han, Tran, Kavuluru, and Mahata]{10.1093/jamia/ocy114}
Gonzalez-Hernandez, G., Sarker, A., Nenadic, G., Belousov, M., Friedrichs, J.,
  Ginter, F., Mehryary, F., Hakala, K., de~Bruijn, B., Mohammad, S.~M.,
  Kiritchenko, S., Rios, A., Han, S., Tran, T., Kavuluru, R., and Mahata, D.
  (2018).
\newblock {Data and systems for medication-related text classification and
  concept normalization from Twitter: insights from the Social Media Mining for
  Health (SMM4H)-2017 shared task}.
\newblock {\em Journal of the American Medical Informatics Association\/}, {\bf
  25}(10), 1274--1283.

\bibitem[Gu {\em et~al.}(2016)Gu, Qian, and Zhou]{gu2016chemical}
Gu, J., Qian, L., and Zhou, G. (2016).
\newblock Chemical-induced disease relation extraction with various linguistic
  features.
\newblock {\em Database\/}, {\bf 2016}.

\bibitem[Karimi {\em et~al.}(2015)Karimi, Metke-Jimenez, Kemp, and
  Wang]{karimi2015cadec}
Karimi, S., Metke-Jimenez, A., Kemp, M., and Wang, C. (2015).
\newblock Cadec: A corpus of adverse drug event annotations.
\newblock {\em Journal of biomedical informatics\/}, {\bf 55}, 73--81.

\bibitem[Klie {\em et~al.}(2018)Klie, Bugert, Boullosa, de~Castilho, and
  Gurevych]{tubiblio106270}
Klie, J.-C., Bugert, M., Boullosa, B., de~Castilho, R.~E., and Gurevych, I.
  (2018).
\newblock The inception platform: Machine-assisted and knowledge-oriented
  interactive annotation.
\newblock In {\em Proceedings of the 27th International Conference on
  Computational Linguistics: System Demonstrations\/}, pages 5--9. Association
  for Computational Linguistics.

\bibitem[Kuratov and Arkhipov(2019)Kuratov and Arkhipov]{kuratov2019adaptation}
Kuratov, Y. and Arkhipov, M. (2019).
\newblock Adaptation of deep bidirectional multilingual transformers for
  russian language.
\newblock {\em arXiv preprint arXiv:1905.07213\/}.

\bibitem[Lee {\em et~al.}(2016)Lee, Hsu, and Kao]{lee2016audis}
Lee, H.-C., Hsu, Y.-Y., and Kao, H.-Y. (2016).
\newblock Audis: an automatic crf-enhanced disease normalization in biomedical
  text.
\newblock {\em Database\/}, {\bf 2016}.

\bibitem[Lee {\em et~al.}(2019)Lee, Yoon, Kim, Kim, Kim, So, and
  Kang]{lee2019biobert}
Lee, J., Yoon, W., Kim, S., Kim, D., Kim, S., So, C.~H., and Kang, J. (2019).
\newblock Biobert: pre-trained biomedical language representation model for
  biomedical text mining.
\newblock {\em Bioinformatics\/}.
\newblock btz682.

\bibitem[Miftahutdinov {\em et~al.}(2017)Miftahutdinov, Tutubalina, and
  Tropsha]{miftahutdinov2017identifying}
Miftahutdinov, Z., Tutubalina, E., and Tropsha, A. (2017).
\newblock Identifying disease-related expressions in reviews using conditional
  random fields.
\newblock In {\em Proceedings of International Conference on Computational
  Linguistics and Intellectual Technologies Dialog\/}, volume~1, pages
  155--166.

\bibitem[Miftahutdinov {\em et~al.}(2020)Miftahutdinov, Alimova, and
  Tutubalina]{ecir2020}
Miftahutdinov, Z., Alimova, I., and Tutubalina, E. (2020).
\newblock On biomedical named entity recognition:experiments in interlingual
  transfer for clinicaland social media texts.
\newblock {\em European Conference on Information Retrieval\/}, {\bf LNCS}.

\bibitem[Moreno {\em et~al.}(2017)Moreno, Boldrini, Moreda, and
  Rom{\'a}-Ferri]{moreno2017drugsemantics}
Moreno, I., Boldrini, E., Moreda, P., and Rom{\'a}-Ferri, M.~T. (2017).
\newblock Drugsemantics: a corpus for named entity recognition in spanish
  summaries of product characteristics.
\newblock {\em Journal of biomedical informatics\/}, {\bf 72}, 8--22.

\bibitem[N{\'e}v{\'e}ol {\em et~al.}(2017)N{\'e}v{\'e}ol, Robert, Anderson,
  Cohen, Grouin, Lavergne, Rey, Rondet, and Zweigenbaum]{neveol2017clef}
N{\'e}v{\'e}ol, A., Robert, A., Anderson, R., Cohen, K.~B., Grouin, C.,
  Lavergne, T., Rey, G., Rondet, C., and Zweigenbaum, P. (2017).
\newblock Clef ehealth 2017 multilingual information extraction task overview:
  Icd10 coding of death certificates in english and french.
\newblock In {\em CLEF (Working Notes)\/}.

\bibitem[N{\'e}v{\'e}ol {\em et~al.}(2018)N{\'e}v{\'e}ol, Robert, Grippo,
  Morgand, Orsi, Pelikan, Ramadier, Rey, and Zweigenbaum]{neveol2018clef}
N{\'e}v{\'e}ol, A., Robert, A., Grippo, F., Morgand, C., Orsi, C., Pelikan, L.,
  Ramadier, L., Rey, G., and Zweigenbaum, P. (2018).
\newblock Clef ehealth 2018 multilingual information extraction task overview:
  Icd10 coding of death certificates in french, hungarian and italian.
\newblock In {\em CLEF (Working Notes)\/}.

\bibitem[Shelmanov {\em et~al.}(2015)Shelmanov, Smirnov, and
  Vishneva]{Shelmanov2015560}
Shelmanov, A., Smirnov, I., and Vishneva, E. (2015).
\newblock Information extraction from clinical texts in russian.
\newblock volume~1, pages 560--572.

\bibitem[Smith {\em et~al.}(2018)Smith, Golder, Sarker, Loke, O’Connor, and
  Gonzalez-Hernandez]{smith2018methods}
Smith, K., Golder, S., Sarker, A., Loke, Y., O’Connor, K., and
  Gonzalez-Hernandez, G. (2018).
\newblock Methods to compare adverse events in twitter to faers, drug
  information databases, and systematic reviews: proof of concept with
  adalimumab.
\newblock {\em Drug safety\/}, {\bf 41}(12), 1397--1410.

\bibitem[Suominen {\em et~al.}(2013)Suominen, Salanter{\"a}, Velupillai,
  Chapman, Savova, Elhadad, Pradhan, South, Mowery, Jones, {\em
  et~al.}]{suominen2013overview}
Suominen, H., Salanter{\"a}, S., Velupillai, S., Chapman, W.~W., Savova, G.,
  Elhadad, N., Pradhan, S., South, B.~R., Mowery, D.~L., Jones, G.~J., {\em
  et~al.} (2013).
\newblock Overview of the share/clef ehealth evaluation lab 2013.
\newblock In {\em International Conference of the Cross-Language Evaluation
  Forum for European Languages\/}, pages 212--231. Springer.

\bibitem[Tutubalina and Nikolenko(2017)Tutubalina and
  Nikolenko]{tutubalina2017combination}
Tutubalina, E. and Nikolenko, S. (2017).
\newblock Combination of deep recurrent neural networks and conditional random
  fields for extracting adverse drug reactions from user reviews.
\newblock {\em Journal of Healthcare Engineering\/}, {\bf 2017}.

\bibitem[Zolnoori {\em et~al.}(2019)Zolnoori, Fung, Patrick, Fontelo, Kharrazi,
  Faiola, Wu, Eldredge, Luo, Conway, {\em et~al.}]{zolnoori2019systematic}
Zolnoori, M., Fung, K.~W., Patrick, T.~B., Fontelo, P., Kharrazi, H., Faiola,
  A., Wu, Y. S.~S., Eldredge, C.~E., Luo, J., Conway, M., {\em et~al.} (2019).
\newblock A systematic approach for developing a corpus of patient reported
  adverse drug events: A case study for ssri and snri medications.
\newblock {\em Journal of biomedical informatics\/}, {\bf 90}, 103091.

\end{thebibliography}

\end{document}